\documentclass[runningheads]{llncs}
\usepackage{graphicx}
\usepackage[misc]{ifsym}

\usepackage{mwe}

\usepackage{xcolor}
\usepackage[table]{xcolor}
\definecolor{lightgray}{gray}{0.9}
\usepackage{colortbl}
\usepackage{amsfonts}
\usepackage{dsfont}
\usepackage{lmodern}
\usepackage{graphicx}
\usepackage{hyperref}
\usepackage[misc]{ifsym}

\makeatletter
\def\blfootnote{\gdef\@thefnmark{}\@footnotetext}
\makeatother

\usepackage{amsthm}

\usepackage{booktabs}
\usepackage{threeparttable}
\usepackage{multirow}
\usepackage{subcaption}

\usepackage{amsmath}
\usepackage{pgfplots}
\pgfplotsset{compat = newest}

\def\relu{\text{ReLU}\xspace}
\def\gelu{\textsc{GELU}\xspace}
\def\gelua{\textsc{GELU}_a\xspace}
\def\iglu{\textsc{IGLU}\xspace}
\def\iglua{\textsc{IGLU-Approx}\xspace}

\def\step{\operatorname{step}\xspace}
\def\igludef{Integrated Gaussian Linear Unit}

\usepackage{multirow}

\usepackage{xcolor,colortbl}
\definecolor{iglugray}{gray}{0.80}  
\begin{document}

\title{IGLU \includegraphics[scale=0.032]{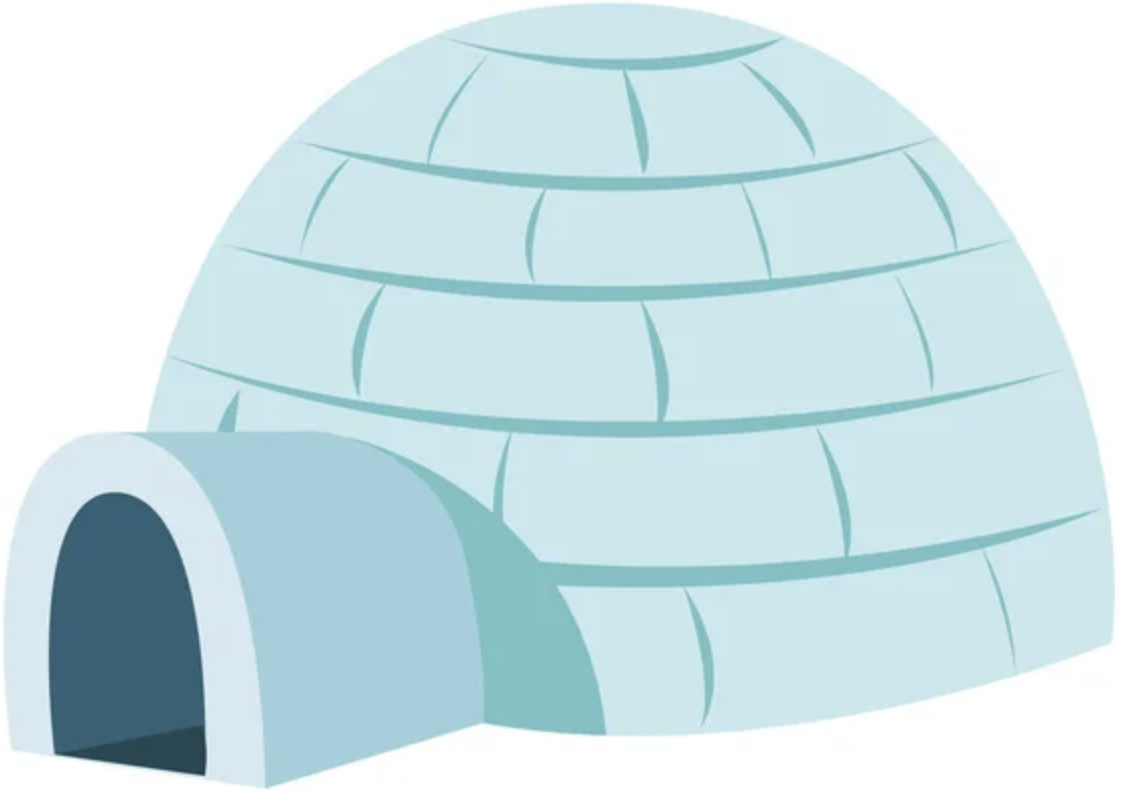}: The Integrated Gaussian Linear Unit Activation Function
\blfootnote{Accepted to ECML PKDD 2026.}
}
\titlerunning{IGLU: The Integrated Gaussian Linear Unit Activation Function}
\toctitle{IGLU: The Integrated Gaussian Linear Unit Activation Function}


\author{Mingi Kang \and Zai Yang \and Jeová Farias Sales Rocha Neto 
}
\authorrunning{M. Kang et al.}
\tocauthor{Mingi Kang, Zai Yang, Jeová Farias Sales Rocha Neto}

\institute{Bowdoin College, 255 Maine Street, Brunswick, ME 04011, USA \\ \email{\{mkang2, zyang, j.farias\}@bowdoin.edu}}

\maketitle 

\begin{abstract}
Activation functions are fundamental to deep neural networks, governing gradient flow, optimization stability, and representational capacity.  Within historic deep architectures, while \relu has been the dominant choice for the activation function, modern transformer-based models increasingly are adopting smoother alternatives such as \gelu and other self-gated alternatives. Despite their empirical success, the mathematical relationships among these functions and the principles underlying their effectiveness remains only partially understood. We introduce \iglu, a parametric activation function derived as a scale mixture of \gelu gates under a half-normal mixing distribution. This derivation yields a closed-form expression whose gating component is exactly the Cauchy CDF, providing a principled one-parameter family that continuously interpolates between identity-like and ReLU-like behavior via a single sharpness parameter $\sigma$. Unlike \gelu's Gaussian gate, \iglu's heavy-tailed Cauchy gate decays polynomially in the negative tail, guaranteeing non-zero gradients for all finite inputs and offering greater robustness to vanishing gradients. We further introduce \iglua, a computationally efficient rational approximation of \iglu expressed entirely in terms of ReLU operations that eliminates transcendental function evaluation. Through evaluations on CIFAR-10, CIFAR-100, and WikiText-103 across ResNet-20, ViT-Tiny, and GPT-2 Small, \iglu achieves competitive or superior performance on both vision and language datasets against \relu and \gelu baselines, with \iglua recovering this performance at substantially reduced computational cost. In particular, we show that employing a heavy-tailed gate leads to considerable performance gains in heavily imbalanced classification datasets. Code for this paper is available at \href{https://github.com/mingikang31/IGLU-Activation}{\url{https://github.com/mingikang31/IGLU-Activation}}. 
\end{abstract}

\section{Introduction}
\label{sec:intro}

Activation functions are a central component of the representational power and optimization dynamics of deep neural networks. By introducing non-linearity between layers, they enable models to approximate complex functions and learn hierarchical structures across a wide range of tasks. While the Rectified Linear Unit (ReLU) \cite{relu} became the dominant activation function in early deep learning systems, its inherent operational issues and the rise of transformer architectures brought smoother alternatives such as the Gaussian Error Linear Unit (GELU) \cite{Hendrycks2016GaussianEL}, Sigmoid Linear Unit (SiLU) \cite{elfwing2018sigmoid} and Mish \cite{misra2019mish}, to mention a few. These smooth activation functions offer improved gradient flow and strong empirical performance, yet the theoretical principles that distinguish them remain only partially understood. 

Despite their widespread adoption, relatively little work has examined how different smooth activation functions relate to one another mathematically, or how systematic modifications to their components influence network behavior. The literature reflects this: of the hundreds of activation functions proposed since the early days of deep learning \cite{kunc2024three}, the vast majority are motivated by empirical intuition rather than principled theoretical reasoning. Existing work largely focuses on benchmarking performance across tasks, with activation design often reduced to heuristic modification of existing functions without a unifying framework. A principled approach, one that grounds activation design in well-understood mathematical structures and offers clear control over the resulting behavior, remains notably uncommon in the literature.

In this work, we introduce \iglu, a novel parametric activation function derived from a continuous mixture of \gelu activations, leading to an arctangent-based function \cite{1628884}. By this construction, \iglu provides a flexible formulation that bridges ReLU and GELU behavior through a single tunable parameter $\sigma$. It further offers a novel connection between neural network training and heavy-tailed data modeling, as it's artangent-based design directly implies the use of a Cauchy CDF gate as an alternative to the Gaussian-one found in \gelu. In practice, due to its slower decay, this formulation guarantees larger gradient magnitudes than \gelu at the activation, a property that improves robustness against vanishing gradients. Finally, inspired by \gelu's more computationally efficient $\operatorname{tanh}(\cdot)$-based approximation, we propose rational approximation of \iglu that does not require the use of any transcendental function, leading to a faster gating mechanism. 

We provide the first systematic evaluation of \iglu on image classification and language modeling tasks, examining its mathematical properties, proposing an efficient approximation suitable for practical deployment, and comparing its empirical performance against ReLU, GELU, and other mainstream activation functions across standard benchmarks. We also provide speed tests that demonstrate \iglu's computational superiority against other common activations and also show how the heavy-tailed modeling within \iglu's formulation contributes better performances at highly unbalanced datasets. Through both algorithmic analysis and experimental validation, we demonstrate \iglu's advantages in representational flexibility, optimization stability, and overall model performance.

\section{\iglu Formulation}
\label{sec:formulation}

Activation functions enable neural networks to learn non-linear representations, and their mathematical structures determine gradient flow, optimization stability, and representational capacity. In this section, we formulate the proposed \iglu activation function and demonstrate how it closely relates to \relu and \gelu. We then analyze the theoretical implications of our proposed activations. Finally, we derive a rational approximation to \iglu and offer a further connection between \relu and itself.  

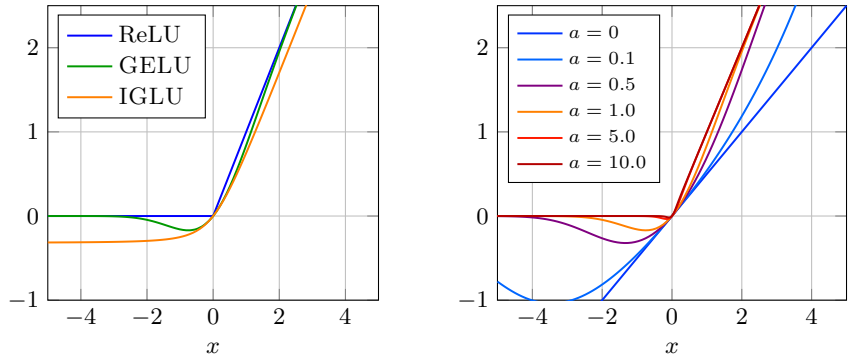
\begin{figure}[t]
\centering
\begin{subfigure}[t]{0.49\textwidth}  
    \centering
    \begin{tikzpicture}
        \begin{axis}
        [
        xmin = -5, xmax = 5, 
        ymin = -1, ymax = 2.5,
        xlabel = {$x$},
        legend pos = north west,
        legend style = {
            font=\footnotesize, cells={anchor=west}
        },
        grid = major,
        width = \textwidth,  
        height = 5.5cm
        ] 
        
        \addplot[
            domain = -5:5,
            samples = 300,
            line width = .75pt,
            blue
        ]{max(0, x)};
        \addlegendentry{\relu}
    
        \addplot[smooth,
            domain = -5:5,
            samples = 50,
            line width = .75pt,
            green!60!black
        ]{x * 0.5 * (1 + tanh(0.797885 * (x + 0.044715 * x^3)))};
        \addlegendentry{\gelu}
        
        \addplot[smooth,
            domain = -5:5,
            samples = 50,
            line width = .75pt,
            orange
        ]{x * (0.5 + atan(x)/180)};
        \addlegendentry{\iglu}
        
        \end{axis}
    \end{tikzpicture}
    \label{fig:relu_gelu_iglu}  
\end{subfigure}
\hfill
\begin{subfigure}[t]{0.51\textwidth}  
    \centering
    \begin{tikzpicture}
        \begin{axis}
        [
        xmin = -5, xmax = 5, 
        ymin = -1, ymax = 2.5,
        xlabel = {$x$},
        legend pos = north west,
        legend style = {
            font=\scriptsize, cells={anchor=west}
        },
        grid = major,
        width = \textwidth,  
        height = 5.5cm
        ] 
        
        \addplot[
            domain = -5:5,
            samples = 10,
            line width = .75pt,
            blue!80!cyan
        ]{x * 0.5 * (1 + tanh(0.797885 * (0 * x + 0.044715 * 0 * x^3)))};
        \addlegendentry{$a = 0$}
            
        \addplot[
            domain = -5:5,
            samples = 100,
            line width = .75pt,
            blue!60!cyan
        ]{x * 0.5 * (1 + tanh(0.797885 * (0.1 * x + 0.044715 * 0.1 * x^3)))};
        \addlegendentry{$a = 0.1$}
        
        \addplot[
            domain = -5:5,
            samples = 100,
            line width = .75pt,
            violet
        ]{x * 0.5 * (1 + tanh(0.797885 * (0.5 * x + 0.044715 * 0.5 * x^3)))};
        \addlegendentry{$a = 0.5$}
        
        \addplot[
            domain = -5:5,
            samples = 100,
            line width = .75pt,
            orange
        ]{x * 0.5 * (1 + tanh(0.797885 * (1 * x + 0.044715 * 1 * x^3)))};
        \addlegendentry{$a = 1.0$}
        
        \addplot[
            domain = -5:5,
            samples = 300,
            line width = .75pt,
            red!80!orange
        ]{x * 0.5 * (1 + tanh(0.797885 * (5.0 * x + 0.044715 * 5.0 * x^3)))};
        \addlegendentry{$a = 5.0$}
        
        \addplot[
            domain = -5:5,
            samples = 300,
            line width = .75pt,
            red!70!black
        ]{x * 0.5 * (1 + tanh(0.797885 * (10.0 * x + 0.044715 * 10.0 * x^3)))};
        \addlegendentry{$a = 10.0$}
        
        \end{axis}
    \end{tikzpicture}
    \label{fig:gelu_a}
\end{subfigure}
\caption{Relevant activation functions in this work. (Left) Comparison of standard activation functions including our proposed \iglu ($\sigma=1$). (Right) Parameterized $\gelua$ for various $a$ values showing convergence to \relu as $a \to \infty$.}\label{fig:relev_act}
\end{figure}

\subsection{Background}

The Rectified Linear Unit (\relu) \cite{relu} is a piecewise linear activation function that is non-smooth at the origin and outputs zero for all negative inputs:
\begin{equation}
    \operatorname{\relu}(x) = \max(0, x) = x \cdot \step(x),
\end{equation}
where $\step(x)$ equals $0$ if $x < 0$ and $1$ otherwise. This step function acts as a hard gating mechanism, determining which activations propagate forward through the network.

\relu is computationally efficient and easy to optimize, but it suffers from the ``dying \relu'' problem due to its zero-gradient region for negative inputs. A widely used smooth alternative is the Gaussian Error Linear Unit (\gelu) \cite{Hendrycks2016GaussianEL}, which replaces the non-differentiable $\operatorname{step}(\cdot)$ in \relu
with a smooth, monotonically increasing function that approximates it. It introduces a soft probabilistic gate based on the Gaussian cumulative distribution function (CDF):
\begin{equation}\label{eq:gelu}
    \gelu(x) = x \cdot \Phi(x), 
    \quad 
    \Phi(x) = \int_{-\infty}^{x} \frac{1}{\sqrt{2\pi}} e^{-t^2/2} dt.
\end{equation}

Unlike \relu, which deterministically suppresses negative inputs, \gelu scales inputs according to their probability under a standard normal distribution, yielding a smooth transition between attenuation and amplification, as shown in Figure~\ref{fig:relev_act} (left). \gelu also carries a probabilistic interpretation as it can be understood as applying a stochastic regularizer to the identity mapping: inputs are retained proportionally to the likelihood that a standard normal random variable falls below $x$ \cite{Hendrycks2016GaussianEL}. This perspective connects \gelu to noise-injection and dropout-like mechanisms, but in a deterministic and differentiable form.

\gelu\ has been widely adopted in transformer architectures such as BERT \cite{devlin2019bert} and GPT \cite{brown2020language} due to its smoother gradient behavior and improved optimization stability in large-scale models. In practice, the exact formulation in Equation~\eqref{eq:gelu} is often approximated for computational efficiency using the following closed-form expression:
\begin{equation}\label{eq:gelu_approx}
\gelu(x) \approx 
\frac{x}{2} + \frac{x}{2}\tanh\!\left(\!
\sqrt{\frac{2}{\pi}}
\left(x + 0.044715 x^3\right)\!
\right),
\end{equation}
which avoids the explicit computation of the Gaussian CDF while closely matching its shape.

Despite its empirical success, \gelu presents certain disadvantages. First, it is computationally more expensive than \relu due to the evaluation of transcendental functions (e.g., $\tanh$, $\operatorname{erf}$). Second, its probabilistic interpretation is tied specifically to the Gaussian distribution, which may not always align with the true statistics of intermediate representations. Finally, the Gaussian gate decays super-exponentially in the negative tail, meaning that strongly negative inputs are suppressed to near zero and their gradients become negligible. This behavior leaves \gelu\ vulnerable to vanishing gradients for large negative pre-activations, and offers no guarantee that any given neuron retains a meaningful gradient signal throughout training.

\subsection{\igludef}

Recent work has explored whether modifying the gating behavior yields performance improvements or better control over gradient dynamics during training. Since \gelu follows the gated-linear form $x \cdot g(x)$, we can introduce a configurable parameter $a$ that scales or sharpens the gate, producing parameterized variants: 
\begin{equation}
    \gelua(x;a) = x \cdot \Phi(a x),
\end{equation}
where $a \geq 0$. As depicted in Figure~\ref{fig:relev_act} (right), adjusting $a$ effectively changes how aggressively the activation function suppresses or amplifies inputs, allowing interpolation between smoother and sharper non-linearities. In both formulations, as $a$ approaches infinity, the function converge to \relu, and as $a$ approaches zero, the function converge to a scaled identity function. At $a = 1$, it reverts back to $\gelu$.

Rather than committing to a single sharpness level $a$, it is natural to ask whether one can aggregate gates of different sharpness into a single adaptive nonlinearity. Instead of fixing $a$, we treat it as a latent scale variable and average over a continuum of gating strengths:
\begin{equation}
    \iglu(x; \sigma)
    =
    \int_{0}^{\infty}
    \gelua(x;a)\, f(a;\sigma)\, da,
\end{equation}
where $f(a;\sigma)$ is a non-negative weighting function parameterized by $\sigma > 0$. The parameter $\sigma$ controls the dispersion of the mixing distribution and therefore governs the effective sharpness of the resulting activation. With \iglu, we hope to promote a scale-mixture interpretation of the $\gelu$ gate, where the resulting activation inherits a multi-scale behavior, combining smooth and sharp nonlinearities in a principled manner. 

Now, when $f(a, \sigma)$ is set to be the half-normal distribution with mean zero and standard deviation $\sigma$, our main challenge is to solve this integral:
\begin{equation}\label{eq:z}
    Z(x;\sigma)= \int_{0}^{\infty} \Phi(a x)\,\frac{2}{\sigma\sqrt{2\pi}}\,
e^{-\frac{a^{2}}{2\sigma^{2}}}\,da,
\end{equation}
where $\iglu(x;\sigma) = x \cdot Z(x, \sigma)$. To solve the integral in Equation~\ref{eq:z}  we can proceed as follows. First we expand $\Phi(ax)$ and substitute $t = as$ (see Equation~\ref{eq:gelu}):
\begin{equation*}
    Z(x;\sigma) = \frac{2x}{\sqrt{2\pi}\,\sigma} \int_0^\infty e^{-a^2/2\sigma^2}
    \int_{-\infty}^{x} \frac{a}{\sqrt{2\pi}}\, e^{-a^2 s^2/2}\, ds\, da.
\end{equation*}

We then swap the order of integration and evaluate the inner integral using
\cite[Formula 3.461.3]{gradshteyn2014table},
\begin{equation*}
    \int_0^\infty a\, e^{-a^2\left(\frac{1}{\sigma^{2}} + s^2\right)/2} da
    = \frac{1}{\sigma^{-2} + s^2} = \frac{\sigma^2}{1 + \sigma^2 s^2},
\end{equation*}
which leads to:
\begin{equation*}
    Z(x;\sigma) = \frac{\sigma x}{\pi} \int_{-\infty}^{x} \frac{ds}{1 + \sigma^2 s^2} = \frac{1}{2} + \frac{\arctan(\sigma x)}{\pi},
\end{equation*}
where we used the result in \cite[Formula 2.01 15] {gradshteyn2014table}. Note that $Z(x, \sigma)$ naturally functions as a smooth gating function replacement for $\step(x)$ in \relu, since it is monotonically increasing ($Z'(x;\sigma) = \sigma / [\pi(1 + \sigma^2 x^2)] > 0$), and
\begin{equation}
    \lim_{x \to -\infty}Z(x;\sigma) =  0, \quad     \lim_{x \to \infty}Z(x;\sigma) =  1.
\end{equation}

Finally, our final expression for the \iglu activation becomes:
\begin{equation}
   \iglu(x; \sigma) 
    = x\left(\frac{1}{2} + \frac{\arctan(\sigma x)}{\pi}\right).
\end{equation}
\vspace{-20pt}

\begin{figure}[t!]
\begin{minipage}[t]{0.43\textwidth}
    \centering
    \begin{tikzpicture}
        \begin{axis}
        [
        xmin = -5, xmax = 5, 
        ymin = -1, ymax = 3,
        xlabel = {$x$},
        legend pos = north west,
        legend style = {
            font=\scriptsize, cells={anchor=west}
        },
        grid = major,
        width = 1.1\textwidth,  
        height = 5.5cm
        ] 
        
        \addplot[
            domain = -5:5,
            samples = 100,
            line width = .75pt,
            blue!80!cyan
        ]{x * (0.5 + atan(0.0*x)/180)};
        \addlegendentry{$\sigma = 0$}
            
        \addplot[
            domain = -5:5,
            samples = 100,
            line width = .75pt,
            blue!60!cyan
        ]{x * (0.5 + atan(0.1*x)/180)};
        \addlegendentry{$\sigma = 0.1$}
        
        \addplot[
            domain = -5:5,
            samples = 100,
            line width = .75pt,
            violet
        ]{x * (0.5 + atan(0.5*x)/180)};
        \addlegendentry{$\sigma = 0.5$}
        
        \addplot[
            domain = -5:5,
            samples = 100,
            line width = .75pt,
            orange
        ]{x * (0.5 + atan(1.0*x)/180)};
        \addlegendentry{$\sigma = 1.0$}
        
        \addplot[
            domain = -5:5,
            samples = 100,
            line width = .75pt,
            red!80!orange
        ]{x * (0.5 + atan(5.0*x)/180)};
        \addlegendentry{$\sigma = 5.0$}
        
        \addplot[
            domain = -5:5,
            samples = 100,
            line width = .75pt,
            red!70!black
        ]{x * (0.5 + atan(10.0*x)/180)};
        \addlegendentry{$\sigma = 10.0$}
        \end{axis}
    \end{tikzpicture}
    \caption{\iglu activation with $\sigma \in \{0.0, 0.1, 0.5, 1.0, 5.0, 10.0\}$. Higher values for $\sigma$ produce sharper transitions at $x =0$.}
    \label{fig:iglu}  
\end{minipage}
\hfill
\begin{minipage}[t]{0.53\textwidth}
    \centering
    \begin{tikzpicture}
        \begin{axis}[
            xmin = -30, xmax = 30, 
            ymin = -0.036, ymax = 0.03,
            xlabel = {$x$},
            legend pos = north west,
            legend style={
                font=\scriptsize,
                at={(0.5, 0.15)},   
                anchor=north,
            },
            legend columns=-1,      
            grid = major,
            width = 1.13\textwidth,  
            height = 5.5cm
        ] 
        
        \addplot[smooth,
            domain = -30:30,
            samples = 100,
            line width = .75pt,
            blue!80!cyan
        ]{((0.5 + atan(0.1*x)/180)) - (((0.5 + max(0, 0.1*x)) / (1 + abs(0.1*x))))};  
        \addlegendentry{$\sigma = 0.1$}
        
        \addplot[smooth,
            domain = -30:30,
            samples = 300,
            line width = .75pt,
            orange
        ]{((0.5 + atan(x)/180)) - (((0.5 + max(0, x)) / (1 + abs(x))))};  
        \addlegendentry{$\sigma = 1.0$}
    
        \addplot[smooth,
            domain = -30:30,
            samples = 1000,
            line width = .75pt,
            red!70!black
        ]{((0.5 + atan(10*x)/180)) - (((0.5 + max(0, 10*x)) / (1 + abs(10*x))))};  
        \addlegendentry{$\sigma = 10.0$}

        \addplot[
            domain = -30:30,
            samples = 2,
            line width = .75pt,
            red, dashed  
        ]{0.023};       
    
        \addplot[
            domain = -30:30,
            samples = 2,
            line width = .75pt,
            red, dashed
        ]{-0.023};
        \end{axis}
    \end{tikzpicture}
    \caption{The difference between $Z(x;\sigma)$ and $Z_{\text{approx}}(x;\sigma)$ for $3$ $\sigma$ values. Maximum absolute difference is never higher than $0.025$ for any $\sigma$.}
    \label{fig:arctan_diff}
\end{minipage}
\end{figure}

\subsection{Algorithmic Implication and Comparison to \gelu}\label{sec:theo}

A central limitation of \relu is that it maps all negative inputs to zero, producing zero gradients in that region and risking neuron death during training. \gelu ameliorates this by replacing the hard step gate with $\Phi(x)$, but the Gaussian CDF itself decays super-exponentially in the negative tail: $\Phi(x) \sim |x|^{-1} e^{-x^2/2}$ as $x \to -\infty$, so large negative inputs are still suppressed to near zero and their gradients become negligible. \iglu replaces this gate with $Z(x;\sigma) = \frac{1}{2} + \frac{\arctan(\sigma x)}{\pi}$, which is precisely the CDF of a Cauchy distribution with scale parameter $\gamma = 1/\sigma$ \cite{andrews1974scale}. The corresponding Cauchy density decays only polynomially as $Z'(x;\sigma) \sim (\sigma \pi x^2)^{-1}$ for large $|x|$, in sharp contrast to the super-exponential decay of the Gaussian gate. This polynomial decay means that even strongly negative inputs retain non-trivial gradients, making \iglu more robust to the vanishing gradient problem \cite{bengio1994learning} relative to both \relu and \gelu. Moreover, since $Z(x;\sigma) > 0$ for all finite $x \in \mathbb{R}$, \iglu guarantees that no neuron can become fully dead during training, a strictly stronger property than \gelu provides and one that \relu cannot offer at all. As pointed out in prior work \cite{Clevert2015FastAA}, this assignment of non-zero outputs to negative inputs allows \iglu to push the mean unit activations closer to zero in a manner analogous to batch normalization, but at lower computational cost. Importantly, this property is not introduced heuristically, as it is the case in \cite{Clevert2015FastAA,DBLP:journals/corr/Barron17a}, but arises naturally from the continuous mixture of \gelu gates: due to the half-Gaussian weighting strategy, most mixture weight is concentrated around \gelu variants that already assign non-zero values to negative entries, as seen in Figure~\ref{fig:relev_act}.

The identification of $Z(x;\sigma)$ as a Cauchy CDF also gives a precise probabilistic interpretation to the scale parameter $\sigma$: it controls the tail weight of the gate's underlying distribution. Smaller values of $\sigma$ correspond to a heavier-tailed Cauchy distribution (scale $\gamma = 1/\sigma$ large), yielding a softer gate that attenuates negative inputs less aggressively; larger values of $\sigma$ correspond to a lighter-tailed distribution and a sharper transition that more closely resembles the hard threshold of \relu. As shown in Figure~\ref{fig:iglu}, this provides a principled one-parameter family interpolating between an identity-like mapping ($\sigma \to 0$) and a hard threshold gate ($\sigma \to \infty$), where the appropriate choice can be matched to the tail behaviour of the network's pre-activations.

This distributional matching argument is particularly relevant when intermediate representations follow heavy-tailed distributions. In such settings, a Gaussian gate systematically under-weights tail inputs relative to their true frequency, whereas the Cauchy-derived gate of \iglu shares the same polynomial tail structure and therefore assigns larger gate values to extreme inputs rather than suppressing them. In this sense, $\sigma$ acts as a distributional matching parameter: it is set to be large when pre-activations are tightly concentrated, and small when they are heavy-tailed. This reasoning is further supported by empirical evidence showing that stochastic gradient noise in deep networks is highly non-Gaussian \cite{simsekli2019tail} and is better characterized by heavy-tailed distributions across a wide range of architectures and datasets. This has a direct implication for activation design: a Gaussian gate is calibrated under the implicit assumption that the signals it operates on are Gaussian-distributed, yet the actual gradient statistics systematically violate this assumption. \iglu's Cauchy gate, by contrast, shares the same polynomial tail structure as $\alpha$-stable distributions (the Cauchy is itself a special case with tail-index $\alpha = 1$) and therefore responds more proportionately to the extreme inputs that arise naturally in heavy-tailed gradient regimes, positioning \iglu as a principled activation choice for the true distributional environment in which deep networks operate.

\subsection{\iglu\ Approximation}

While the formulation above provides smooth and consistent gradients, evaluating the $\arctan(\cdot)$ function introduces non-negligible computational overhead in deep networks, particularly in large-scale models where activation functions are applied billions of times during training. To address this limitation, we propose \iglua, a lightweight rational approximation that preserves the qualitative behavior of the original $Z(x;\sigma)$ gate while relying only on elementary operations.

We begin by employing the following approximation for the arctangent function:
\begin{equation}\label{eq:a_approx}
   \arctan(\sigma x) \approx \frac{\pi}{2}\frac{\sigma x}{1+|\sigma x|},
\end{equation}
which is continuous, odd, and saturates correctly as $x \to \pm\infty$. Substituting~\eqref{eq:a_approx} into the original gating function yields
\begin{equation}
    Z_{\text{approx}}(x;\sigma)
    =
    \frac{1}{2}
    \frac{1 + 2\max(0,\sigma x)}
    {1 + |\sigma x|} =
    \frac{1}{2}
    \frac{1 + 2\relu(\sigma x)}
    {1 + \relu(\sigma x) + \relu(-\sigma x)},
\end{equation}
where we used the identities $x + |x| = 2\max(0,x)$ and $|x| = \max(0,x) + \max(0,-x)$. Importantly, note that this approximation admits a formulation entirely in terms of \relu activations.

Figure~\ref{fig:arctan_diff} empirically shows that the maximum deviation between the original and approximated gates does not exceed $0.05$ for any $x$, and that the approximation error vanishes as $x \to \pm\infty$. This behavior holds uniformly for all $\sigma \geq 0$, indicating that the approximation preserves both the global shape and asymptotic structure of the original gate.

Our final activation, $\iglua = x \cdot Z_{\text{approx}}(x;\sigma)$, is then defined as
\begin{equation}
     \iglua(x;\sigma)
     =
    \frac{x}{2}
    \frac{1 + 2\relu(\sigma x)}
    {1 + \relu(\sigma x) + \relu(-\sigma x)}.
\end{equation}

By combining heavy-tailed gating behavior with a rational, \relu-compatible formulation, \iglua\ achieves a principled trade-off between expressiveness and computational efficiency. In contrast to \gelu, whose practical implementation (cf.~Equation~\ref{eq:gelu_approx}) still depends on transcendental functions, \iglua\ is expressed exclusively through ReLU operations and basic arithmetic. This eliminates costly special-function evaluations while preserving smooth, saturating gating dynamics. As a result, \iglua\ is particularly well suited for large-scale architectures and resource-constrained environments, where activation efficiency, numerical stability, and hardware-friendly execution are critical.

\section{Related Works}
The arctangent function has appeared in the neural network literature as a smooth alternative to piecewise-linear activations. Lederer~\cite{lederer2021activation} provides a systematic overview of activation functions and discusses the mathematical properties that make $\arctan(\cdot)$ appealing, including smoothness, boundedness, and odd symmetry. Similarly, Sivri et al.~\cite{sivri2022multiclass} investigates multiclass classification using arctangent-based activations, leveraging the facts of $\arctan(x) \in (-\frac{\pi}{2}, \frac{\pi}{2})$ and $\arctan(-x) = -\arctan(x)$. These works establish $\arctan(\cdot)$ as a viable smooth activation; however, they consider it in its raw, non-normalized form, bounded in $(-\frac{\pi}{2}, \frac{\pi}{2})$, and do not situate it within the multiplicative gating framework that characterizes modern activations such as \gelu\ and SiLU. In contrast, our formulation employs a normalized version of $\arctan(\cdot)$, mapping it to the interval $(0,1)$, which arises naturally from our scale-mixture interpretation of the \gelu\ gate~\cite{andrews1974scale}. This normalization is not ad hoc, but is instead theoretically motivated by its connection to cumulative distribution functions.

Zhang et al.~\cite{zhang2024deep} study deep network approximation beyond \relu, explicitly analyzing activations of the form $f(x)=x\cdot g(x)$, where $g(x)$ may include the (unnormalized) arctangent. Their theoretical results provide approximation error bounds and demonstrate that arctangent-based gates can match the expressive power of \relu. Nevertheless, their analysis does not consider the normalized $(0,1)$ version of $\arctan$, does not introduce a principled sharpness parameter governing the transition between linear and rectified regimes, and does not examine the relationship between such gates and \gelu\ in practical training. Moreover, the connection between normalized $\arctan(\cdot)$ gating and the Cauchy cumulative distribution function is not identified or explored, thereby overlooking the implications of heavy-tailed modeling for representation learning.

The work most closely related to ours is the recent preprint by Huang~\cite{huang2024expanded}, which proposes Expanding Gating Ranges (xATLU), extending the ArcTan Linear Unit defined as $\operatorname{ATLU}(x) = x \left( \frac{\arctan(x) + \frac{\pi}{2}}{\pi} \right)$.
The xATLU propsed extension introduces a trainable scalar $\alpha$ that expands the gating range from $(0,1)$ to $(-\alpha, 1+\alpha)$: $
\mathrm{xATLU}(x,\alpha) = x \left( (1+2\alpha)\frac{\arctan(x)+\frac{\pi}{2}}{\pi} - \alpha \right)$. Huang demonstrates that learning the gating range can improve empirical performance across ATLU, \gelu, and SiLU variants. Although xATLU and \iglu\ share an arctangent-based gating structure, several fundamental differences remain. First, xATLU modifies the output range through affine scaling, whereas \iglu\ modulates the input sharpness via the parameter $\sigma$, directly controlling the transition between linear and \relu-like behavior. Second, xATLU learns $\alpha$ during training, while \iglu\ employs a fixed hyperparameter $\sigma$ with a clear statistical interpretation: it governs the dispersion of the underlying Cauchy mixing distribution and thus adapts the activation to light- or heavy-tailed regimes. Third, \iglu\ provides an explicit connection between \relu and \gelu, together with a principled derivation that demonstrates convergence to \relu\ as $\sigma \to \infty$. This derivation also naturally motivates the normalization of $\arctan$ to $(0,1)$ via its identification with the Cauchy CDF, rather than introducing the normalization heuristically. Finally, neither xATLU nor prior arctangent-based activations develop the connection to heavy-tailed modeling or analyze the implications for learning under long-tailed data distributions.

An additional distinguishing contribution of our work is the introduction of the \iglua\ approximation. The arctangent-based works rely on direct evaluation of $\arctan(\cdot)$, which entails transcendental function computation. In contrast, we derive a lightweight rational approximation that replaces all transcendental evaluations with elementary arithmetic and \relu\ operations. This significantly improves computational efficiency while preserving the qualitative gating behavior, making \iglua\ particularly suitable for large-scale and resource-constrained training environments. To our knowledge, no prior work provides a comparable rational approximation within an arctangent-based multiplicative gating framework.

\section{Experiments and Discussion}
\label{sec:experiments}

\begin{table}[t!]
\centering
\small
\setlength{\tabcolsep}{9pt}
\caption{Computational efficiency benchmarks for various activation functions. Forward and backward pass times are normalized relatively to the Identity function speed. Results are averaged over 1,000 iterations using a 10,000-dimensional input. Gray shading denotes self-gating activations of the form $f(x) = x \cdot g(x)$. \textbf{Bold} and \underline{underline} indicate the best and second-best performance within the self-gating category, respectively.}  

\label{tab:speed}
\vspace{5pt}
\begin{tabular}{lccccc}
\toprule
&& \multicolumn{2}{c}{CPU}  & \multicolumn{2}{c}{GPU (CUDA)} \\
\cmidrule(lr){3-4} \cmidrule(lr){5-6}
\multicolumn{2}{l}{Activation} &  Forward & Backward & Forward & Backward\\
\midrule[1pt]
$Z(\cdot)$ ($\sigma = 1$)                                     & & 21.95$\times$ & 3.43$\times$ & 14.77$\times$ & 2.80$\times$ \\
$Z_{\text{approx}}(\cdot)$  ($\sigma = 1$)                                   & & 10.13$\times$ & 3.61$\times$ & 15.56$\times$ & 3.14$\times$ \\
\rowcolor{lightgray} \iglu  ($\sigma = 1$)                                    & & 22.44$\times$ & 4.39$\times$ & 14.89$\times$ & 2.68$\times$ \\
\rowcolor{lightgray} \iglua  ($\sigma = 1$)                                      & & 10.17$\times$ & 3.63$\times$ & 15.05$\times$ & \textbf{2.23}$\times$ \\
\midrule[0.5pt]
\rowcolor{lightgray} ReLU \cite{relu}   & & \textbf{9.22}$\times$  & \textbf{3.39}$\times$ & \underline{14.85}$\times$ & 2.66$\times$ \\
\rowcolor{lightgray} GELU \cite{Hendrycks2016GaussianEL}         & & 16.40$\times$ & 4.32$\times$ & 15.36$\times$ & \underline{2.27}$\times$ \\
\rowcolor{lightgray} SiLU \cite{elfwing2018sigmoid}              & & 13.61$\times$ & 3.97$\times$ & 15.31$\times$ & 3.04$\times$ \\
\midrule[0.5pt]
CELU \cite{DBLP:journals/corr/Barron17a}         & & 14.63$\times$ & 3.63$\times$ & 15.60$\times$ & 2.88$\times$ \\
ELU \cite{Clevert2015FastAA}                   & & 13.81$\times$ & 3.71$\times$ & 14.85$\times$ & 3.18$\times$ \\
Hardshrink \cite{donoho1995noising}             & & 9.50$\times$  & 3.42$\times$ & 15.58$\times$ & 3.04$\times$ \\
Hardsigmoid \cite{howard2019searching}      & & 9.54$\times$  & 3.32$\times$ & 15.44$\times$ & 2.41$\times$ \\
\rowcolor{lightgray} Hardswish \cite{howard2019searching}        & & \underline{9.71}$\times$  & \underline{3.43}$\times$ & 16.02$\times$ & 2.52$\times$ \\
HardTanh \cite{lupu2024exact}                   & & 9.58$\times$  & 3.34$\times$ & 16.74$\times$ & 3.32$\times$ \\
LeakyReLU \cite{maas2013rectifier}              & & 9.50$\times$  & 4.00$\times$ & 14.73$\times$ & 3.18$\times$ \\
\rowcolor{lightgray} Mish \cite{misra2019mish}                   & & 64.08$\times$ & 7.77$\times$ & \textbf{14.75}$\times$ & 2.39$\times$ \\
PReLU \cite{he2015delving}                      & & 9.86$\times$  & 11.00$\times$& 18.34$\times$ & 3.39$\times$ \\
SELU \cite{klambauer2017self}                   & & 14.39$\times$ & 3.71$\times$ & 14.76$\times$ & 3.53$\times$ \\
Sigmoid                                   & & 13.47$\times$ & 3.43$\times$ & 14.79$\times$ & 3.01$\times$ \\
Softplus \cite{dugas2000incorporating}          & & 28.82$\times$ & 3.72$\times$ & 14.74$\times$ & 3.22$\times$ \\
Softshrink \cite{donoho1995noising}             & & 10.11$\times$ & 3.49$\times$ & 16.42$\times$ & 2.77$\times$ \\
Squareplus \cite{barron2021squareplus}      & & 10.72$\times$ & 3.52$\times$ & 14.90$\times$ & 2.20$\times$ \\
Tanh                                            & & 26.78$\times$ & 3.50$\times$ & 15.76$\times$ & 2.82$\times$ \\
Tanhshrink \cite{donoho1995noising}             & & 27.02$\times$ & 4.73$\times$ & 15.05$\times$ & 2.99$\times$ \\
\midrule[0.5pt]
Identity \cite{he2016identity}                  & & 1.00$\times$  & 1.00$\times$ & 1.00$\times$  & 1.00$\times$ \\

\bottomrule[1pt]
\end{tabular}
\end{table}

\begin{table}[t]
\centering
\caption{Test loss and test accuracy of ResNet-20 and ViT-Tiny trained on CIFAR-10 and CIFAR-100. \textbf{Bold} and \underline{underline} indicate the best and second-best performance, respectively. ViT-Tiny experiments undergo a resizing of size $224 \times 224$ to fit the Vision Transformer training procedure.}
\label{tab:resnet_vit_baseline}
\vspace{5pt}    

\resizebox{\linewidth}{!}{
    \begin{tabular}{lcccccccc} 
    \toprule
    & \multicolumn{4}{c}{ResNet-20} & \multicolumn{4}{c}{ViT-Tiny} \\
    \cmidrule(lr){2-5} \cmidrule(lr){6-9}
    & \multicolumn{2}{c}{CIFAR-10}  & \multicolumn{2}{c}{CIFAR-100} & \multicolumn{2}{c}{CIFAR-10}  & \multicolumn{2}{c}{CIFAR-100}\\
    \cmidrule(lr){2-3} \cmidrule(lr){4-5} \cmidrule(lr){6-7} \cmidrule(lr){8-9} 
    Activation & Test Loss & Acc & Test Loss & Acc & Test Loss & Acc & Test Loss & Acc \\
    \midrule[1pt]
    \iglu ($\sigma = 0.1$)       & \underline{0.389} & 88.59 & \underline{1.377} & 62.51 & 1.251 & 82.81 & 3.769 & 53.60 \\ 
    \iglu ($\sigma = 0.5$)       & 0.480 & \textbf{91.10} & 1.488 & \underline{65.46} & 1.211 & 85.23 & 3.585 & 57.55 \\
    \iglu ($\sigma = 1$)         & 0.522 & 90.94 & 1.560 & 64.64 & 1.088 & 86.65 & 3.240 & \underline{61.83} \\
    \iglu ($\sigma = 5$)         & 0.509 & 90.71 & 1.601 & 63.75 & 1.099 & 86.70 & 3.193 & 61.74 \\
    \iglu ($\sigma = 10$)        & 0.520 & 90.66 & 1.565 & 64.16 & 1.072 & \underline{87.27} & 3.127 & 61.82 \\
    \iglu (learnable $\sigma$)   & 0.546 & 90.39 & 1.561 & 64.78 & 1.085 & 86.72 & 3.137 & 61.77 \\
    \midrule[0.5pt]
    \iglua ($\sigma = 0.1$)      & \textbf{0.374} & 89.39 & \textbf{1.350} & 63.81 & 1.336 & 81.15 & 3.800 & 53.42 \\
    \iglua ($\sigma = 0.5$)      & 0.505 & 90.44 & 1.462 & \textbf{65.88} & 1.284 & 83.76 & 3.628 & 56.70 \\
    \iglua ($\sigma = 1$)        & 0.502 & 90.81 & 1.508 & 65.18 & 1.029 & 86.75 & 3.185 & 61.61 \\
    \iglua ($\sigma = 5$)        & 0.511 & \underline{91.01} & 1.598 & 63.50 & 1.119 & 87.17 & 3.357 & 61.29 \\
    \iglua ($\sigma = 10$)       & 0.508 & 90.60 & 1.573 & 63.97 & 1.070 & 86.80 & 3.182 & 60.91 \\
    \iglua (learnable $\sigma$)  & 0.521 & 90.70 & 1.565 & 64.48 & 1.058 & 87.18 & 3.216 & 61.17 \\
    \midrule[.5pt]
    ReLU                         & 0.487 & 90.69 & 1.558 & 62.28 & 0.993 & 86.98 & \textbf{3.009} & 61.79 \\
    GELU                         & 0.539 & 90.72 & 1.630 & 63.51 & 1.136 & 86.50 & 3.292 & 60.89 \\
    SiLU                         & 0.566 & 90.56 & 1.607 & 64.56 & 1.035 & 87.24 & 3.194 & \textbf{63.10} \\
    Hardswish                    & 0.561 & 90.24 & 1.606 & 64.40 & \textbf{1.008} & 86.66 & \underline{3.126} & 61.74 \\
    Mish                         & 0.586 & 90.08 & 1.611 & 64.61 & \underline{1.018} & \textbf{87.96} & 3.479 & 61.28 \\
    \bottomrule
    \end{tabular}
}
\end{table}

In this section we evaluate \iglu on speed tests and on both vision and language tasks to assess its effectiveness across different domains and architectures. We further explore its performance on highly imbalanced datasets. 

\subsection{Speed tests}

Table~\ref{tab:speed} benchmarks the forward and backward pass times of a range of popular activation functions in both CPU and GPU environments, measured in isolation outside of network training. For completeness, we include our gate $Z(\cdot)$ and its approximation $Z_{\text{approx}}(\cdot)$ alongside the baselines, and use \gelu's closed-form approximation (Equation~\ref{eq:gelu_approx}) throughout. We use these same implementations in the following sections. All experiments in Table \ref{tab:speed} are implemented in PyTorch 2.0 and executed on a 13th Gen Intel Core i9-13900H CPU and a NVIDIA GeForce RTX 4070 PCIe 16GB GPU. Among all activations, \iglua's speed is competitive, matching the speed of functions that rely exclusively on non-transcendental operations such as \relu, Hardshrink, Hardswish and HardTanh, particularly in forward passes. Compared to \gelu's approximation, \iglua is consistently faster across the table. Broadly speaking, the results confirm that \iglua substantially reduces the computational overhead of \iglu, most notably on CPU, bringing it in line with the most efficient activations in the benchmark while preserving the heavy-tailed gating behavior of the original formulation.

\subsection{Vision Models} \label{subsec/Vision_Models}

We evaluate \iglu on two architectures spanning different design paradigms: ResNet-20 \cite{He2015DeepRL}, a classical CNN where activation functions follow convolutional layers, and ViT-Tiny \cite{dosovitskiy2020image}, a Vision Transformer where activations are applied within the MLP blocks of each encoder layer. Both are trained on CIFAR-10 and CIFAR-100 \cite{krizhevsky2009learning}, standard benchmarks of $32 \times 32$ color images across 10 and 100 classes, respectively. ResNet-20 experiments use the native $32 \times 32$ resolution, while ViT-Tiny requires resizing to $224 \times 224$ to accommodate it's patch embedding process.

For all image classification tasks, we apply standard data augmentations during training: random cropping, horizontal flips, and color jittering. No augmentations are applied to the test set to ensure fair evaluation. For all vision models on image classification tasks, we train for 200 epochs using the AdamW optimizer with a learning rate of $1\times10^{-3}$, weight decay of $1 \times 10^{-2}$, and batch size of 128. We apply a cosine annealing learning rate scheduler to gradually reduce the learning rate over training. All experiments are implemented in PyTorch 2.0 and executed on NVIDIA A100 PCIe 80GB GPU.

Table \ref{tab:resnet_vit_baseline} displays \iglu and \iglua's test loss and accuracy performance on both datasets for both networks. We vary the value of $\sigma$ and also test our proposed activations when $\sigma$ is a learnable parameter, comparing them quantitatively against the self gating activations displayed in Table \ref{tab:speed}. While our proposed activations perform well at lower $\sigma$ in ResNet (to the point over-performing all compared activations by a considerable margin), higher $\sigma$ is preferable in ViT. This pattern offers us evidence that the data flowing in the ResNet's convolution layers follows a distribution that is better matched by an activation that expects heavy-tailed data, such as low-$\sigma$ \iglu. In fact, these results suggest that \iglu is the most suited activation among all popular self-gated ones.  On the other hand, possibly due to the presence of layer normalization in ViT, the data flowing through it is better modeled with lighter-tailed activations, given the success of high-$\sigma$ \iglu and \relu. Finally, the learnable $\sigma$ experiment show that having our activations adapt themselves while training proved unsuccessful, potentially hitting at instability caused by the operation itself or overfitting. 

\subsection{Language Models}

\begin{table}[t]
\centering
\small
\setlength{\tabcolsep}{9.5pt}
    \caption{Test loss and test perplexity (the lower the better) of GPT2-Small trained on WikiText-103. \textbf{Bold} and \underline{underline} indicate the best and second-best performance, respectively.}\label{tab:gpt2_baseline}
    \vspace{5pt}    
    \begin{tabular}{lcc}
    \toprule
    Activation &  Test Loss  & Test Perplexity\\
    \midrule[1pt]
    \iglu ($\sigma = 0.1$)      & 2.989 & 16.69 \\
    \iglu ($\sigma = 0.5$)      & 2.814 & 16.69 \\
    \iglu ($\sigma = 1$)        & 2.761 & 15.82 \\
    \iglu ($\sigma = 5$)        & \textbf{2.754} & \textbf{15.71} \\
    \iglu ($\sigma = 10$)       & \underline{2.755} & \underline{15.72} \\
    \iglu (learnable $\sigma$)  & \textbf{2.754} & \textbf{15.71} \\
    \midrule[.5pt]
    \iglua ($\sigma = 0.1$)    & 2.987 & 19.84 \\
    \iglua ($\sigma = 0.5$)    & 2.809 & 16.60 \\
    \iglua ($\sigma = 1$)      & 2.772 & 15.99 \\
    \iglua ($\sigma = 5$)      & 2.756 & 15.74 \\
    \iglua ($\sigma = 10$)     & 2.756 & 15.74 \\
    \iglua (learnable $\sigma$)& 2.756 & 15.74 \\
    \midrule[.5pt]
    ReLU                         & \underline{2.755} & 15.73 \\
    GELU                         & 2.761 & 15.82 \\
    SiLU                         & 2.767 & 15.92 \\
    Hardswish                    & 2.781 & 16.15 \\
    Mish                         & 2.766 & 15.90 \\
    \bottomrule
    \end{tabular}
\end{table}

For language modeling, we train GPT-2 Small \cite{radford2019language} on WikiText-103 \cite{DBLP:journals/corr/MerityXBS16}, a standard benchmark comprising over 100 million tokens drawn from verified Wikipedia articles. We evaluate performance using perplexity, which measures how well the model's predicted distribution aligns with the true word distribution, where lower values indicate better performance. Activation functions in GPT-2 are applied within the feed-forward network of each transformer block, making this a particularly relevant testbed for understanding how activation choice affects autoregressive language modeling at scale. We adopt the standard GPT-2 Small architecture with a vocabulary size of 50,257, maximum sequence length of 1,024, embedding dimension of 768, 12 attention heads, and 12 transformer layers. All models are trained for 20 epochs with a batch size of 32 using the AdamW optimizer with a learning rate of $6\times 10^{-4}$ and weight decay of $0.1$, with a linear learning rate scheduler and a warmup period covering 5\% of total training steps. Experiments are implemented in PyTorch 2.0 and executed on an NVIDIA GH200 GPU with 480 GB of memory and 72 CPU core Neoverse-V2 ARM processor. 

Table~\ref{tab:gpt2_baseline} reports results for \iglu and \iglua against the self-gated baselines from Table~\ref{tab:speed} using the same $\sigma$'s from Table \ref{tab:resnet_vit_baseline}. Both proposed activations perform strongly and consistently at higher values of $\sigma$, while underperforming at lower ones, suggesting that a lighter-tailed gate is better suited to this setting, likely due again to the presence of layer normalization in GPT-2, which pushes pre-activations toward a near-Gaussian regime as discussed in Section~\ref{sec:theo}. Notably, at $\sigma = 5$, \iglu and \iglua surpass not only \gelu and \relu, our motivating activations, but all other self-gated activations in the comparison, providing strong evidence for the suitability of \iglu in large-scale language modeling. The learnable $\sigma$ offers no performance advantage over its fixed counter parts, only reaching parity with the best results. This reinforces previous observations regarding to training instability and overfitting. 

\subsection{Imbalanced Datasets}
\begin{table*}[tb]
\centering
\setlength{\tabcolsep}{2pt}
\caption{Performance of ResNet-20 on the Imbalanced CIFAR-100 dataset across various imbalance ratios (where, for example, $100{:}1$ indicates $n_{\max} / n_{\min} = 100$)). We report the cross-entropy test loss (Loss) and test accuracy (Acc) for our proposed and baseline activation functions. \textbf{Bold} and \underline{underline} indicate the best and second-best performance, respectively.} \label{tab:imbalance}
\vspace{5pt}    
\resizebox{1\linewidth}{!}{
    \begin{tabular}{lcccccccccc}
    \toprule
     & \multicolumn{2}{c}{10:1} & \multicolumn{2}{c}{20:1} & \multicolumn{2}{c}{50:1} & \multicolumn{2}{c}{100:1} & \multicolumn{2}{c}{500:1}\\
    \cmidrule(lr){2-3} \cmidrule(lr){4-5} \cmidrule(lr){6-7} \cmidrule(lr){8-9} \cmidrule(lr){10-11}
    Activation                            & Loss & Acc & Loss & Acc & Loss & Acc & Loss & Acc & Loss & Acc \\
    \midrule
    \iglu ($\sigma = 0.1$)      & \underline{2.078} & 46.69 & \underline{2.456} & 38.60 & \underline{2.982} & 29.43 & \textbf{3.426} & 23.05 & \underline{3.957} & 17.99 \\
    \iglu ($\sigma = 0.5$)      & 2.719 & 49.99 & 3.309 & \underline{44.97} & 3.923 & \textbf{37.76} & 4.878 & 31.45 & 5.659 & \underline{28.94} \\
    \iglu ($\sigma = 1$)        & 2.833 & 50.02 & 3.514 & 43.95 & 4.482 & 36.24 & 5.352 & 31.64 & 6.192 & \textbf{29.03} \\
    \iglu ($\sigma = 5$)        & 2.931 & 49.21 & 3.675 & 43.63 & 4.756 & 35.27 & 5.456 & 30.15 & 6.182 & 28.80 \\
    \iglu ($\sigma = 10$)       & 2.878 & 49.48 & 3.570 & 43.49 & 4.651 & 35.48 & 5.350 & 30.89 & 6.014 & 28.37 \\
    \midrule[.5pt]
    \iglua 
    ($\sigma = 0.1$)    & \textbf{2.043} & 48.37 & \textbf{2.390} & 40.75 & \textbf{2.919} & 32.12 & \underline{3.452} & 23.01 & \textbf{3.938} & 20.60 \\
    \iglua  
    ($\sigma = 0.5$)    & 2.558 & \textbf{51.20} & 3.074 & \textbf{45.82} & 4.023 & 36.29 & 4.784 & \textbf{32.46} & 5.454 & 28.84 \\
    \iglua 
    ($\sigma = 1$)      & 2.763 & \underline{51.06} & 3.458 & 44.49 & 4.558 & \underline{36.51} & 5.154 & \underline{32.43} & 6.089 & 28.13 \\
    \iglua 
    ($\sigma = 5$)     & 2.986 & 49.34 & 3.558 & 43.74 & 4.611 & 35.24 & 5.447 & 31.74 & 6.381 & 28.38 \\
    \iglua  
    ($\sigma = 10$)     & 2.944 & 49.39 & 3.622 & 43.64 & 4.666 & 34.94 & 5.347 & 30.52 & 6.211 & 27.54 \\
    \midrule[.5pt]
    ReLU            & 2.823 & 49.26 & 3.570 & 43.99 & 4.448 & 36.28 & 5.436 & 30.01 & 5.711 & 26.66 \\
    GELU            & 3.048 & 50.28 & 3.751 & 44.64 & 4.705 & 35.08 & 5.603 & 30.17 & 6.237 & 26.77 \\
    SiLU            & 2.988 & 50.45 & 3.737 & 44.15 & 4.933 & 34.81 & 5.693 & 31.31 & 6.377 & 28.18 \\
    Hardswish       & 3.011 & 49.55 & 3.761 & 43.65 & 4.773 & 35.61 & 5.685 & 30.08 & 6.251 & 26.15 \\
    Mish            & 3.028 & 50.17 & 3.648 & 44.56 & 4.619 & 36.34 & 5.578 & 31.83 & 6.132 & 28.79 \\
    \bottomrule
    \end{tabular}
}
\label{fig:imbalanced_baseline}
\end{table*}

Finally, we evaluate \iglu on the long-tailed variant of CIFAR-100, CIFAR-100-LT, constructed following the protocol of \cite{cui2019class}. The long-tailed version is obtained by exponentially subsampling the training data so that the number of samples per class follows 
\begin{equation}
    n_k = n_{\max} \cdot \mu^{\frac{k}{K-1}}, \quad k = 0, 1, \ldots, K-1,
\end{equation}
where $K$ is the total number of classes, $n_{\max}$ is the number of samples in the most frequent class, and $\mu = n_{\min} / n_{\max} \in (0, 1]$ is the \textit{imbalance factor} controlling the severity of the imbalance. The corresponding \textit{imbalance ratio} (IR) is $\rho = 1/\mu = n_{\max} / n_{\min}$. In Table~\ref{tab:imbalance}, the imbalance ratios are denoted as $\rho{:}1$ (e.g., $100{:}1$ indicates $n_{\max} / n_{\min} = 100$). The test set remains balanced with all classes represented equally. 

This construction induces a class-frequency distribution characterized by a small number of well-represented \textit{head} classes and many sparsely sampled \textit{tail} classes. From a statistical perspective, such regimes are reminiscent of heavy-tailed phenomena, where rare but structurally important events dominate estimation dynamics. This heavily imbalanced class distribution creates a high dispersion in images and therefore feature norms, where the representation geometry is dominated by a small subset of frequent classes, mimicking heavy-tail influence phenomena observed in continuous heavy-tailed distributions. Therefore we hypothesize that \iglu with its capacity of retaining large activations may introduces a implicit robustness to class skewness and prove itself as a better methodological neural network component than \relu and \gelu. 

Following the image classification setup described in Section~\ref{subsec/Vision_Models}, we adopt the same training hyperparameters but replace the loss function with class-weighted cross-entropy to account for class imbalance during training. For evaluation, we use standard unweighted cross-entropy on the balanced test set and report accuracy as the primary metric, which weights each class equally regardless of its training frequency. Table~\ref{tab:imbalance} presents results for ResNet-20 trained on CIFAR-100-LT across imbalance ratios $\rho \in \{10, 20, 50, 100, 500\}$, and we again restrict comparisons to self-gated activations for consistency. Two clear trends emerge. First, \iglu and \iglua achieve their strongest performance at lower values of $\sigma$, which we attribute to the heavier-tailed Cauchy gate providing a better distributional match to the skewed input statistics induced by class imbalance, and potentially promoting more stable gradient flow for underrepresented classes. Second, as the value $\sigma$ increases, performance converges toward the \relu baseline, which is consistent with the theoretical analysis in Section~\ref{sec:theo} and further underscores the unsuitability of sharp-gating activations, including \relu (the $a \to \infty$ limit of $\gelua$) and \gelu ($a = 1$), for heavily imbalanced settings.

\section{Pre-Activation Analysis with Empirical Tail Index}

\begin{table}[t]
\centering
\small
\setlength{\tabcolsep}{9.5pt}
    \caption{Empirical Index Index $\alpha$ of pre-activation measurements using the Hill Estimator with and without Normalization (Batch Normalization for ResNet, Layer Normalization for ViT and GPT-2) using \iglu activation function.}
    \label{tab:empirical_tail_index}
    \vspace{5pt}    
    \begin{tabular}{lccc}
    \toprule
    Model & Norm & $\alpha$ & Optimal $\sigma$ \\
    \midrule[1pt]
    ResNet-20  & BatchNorm & 10.4 & 0.1 \\
    GPT2-Small & LayerNorm & 20.7 & 5.0 \\
    ViT-Tiny   & LayerNorm & 23.2 & 5.0 \\
    \midrule[1pt]
    ResNet-20  & - & 11.28 & 0.1 \\
    \bottomrule
    \end{tabular}
\end{table}

In Table~\ref{tab:empirical_tail_index}, we measure the empirical tail index $\alpha$ of pre-activations across three trained architectures using the Hill estimator \cite{10.1214/aos/1176343247}. ResNet-20 with BatchNorm yields $\alpha = 10.4$, while ViT-Tiny and GPT2-Small with LayerNorm yields $\alpha = 23.2$ and $\alpha = 20.7$, respectively. This ordering directly corresponds to the optimal $\sigma$ values identified in Table~\ref{tab:resnet_vit_baseline} and Table~\ref{tab:gpt2_baseline}: the heavier-tailed ResNet pre-activations (low $\alpha$) favors $\sigma = 0.1$, while the lighter-tailed Transformer pre-activations (high $\alpha$) favors $\sigma = 5.0$, validating the distributional matching hypothesis. Notably, removing normalization entirely caused ViT-Tiny and GPT-2 Small to diverge during training, consistent with the known instability of Transformer architecture without LayerNorm and further highlighting the role of normalization in shaping the pre-activation distribution.

\section{Conclusion}
We introduced \iglu, a parametric activation function derived as a continuous scale mixture of \gelu gates, yielding a closed-form gating component that is exactly the Cauchy CDF. This construction provides a principled one-parameter family interpolating between identity-like and \relu-like behavior via a single sharpness parameter $\sigma$, whose polynomial tail decay guarantees non-zero gradients for all finite inputs and offers greater robustness to vanishing gradients than either \relu or \gelu. The heavy-tailed gating structure is further theoretically motivated by empirical evidence that stochastic gradient noise in deep networks follows heavy-tailed $\alpha$-stable distributions \cite{simsekli2019tail}, positioning \iglu as a principled activation choice for the true distributional environment in which deep networks are optimized. We additionally introduced \iglua, a rational approximation expressed entirely in terms of \relu operations that eliminates transcendental function evaluation while preserving the qualitative gating behavior of the original formulation. Across image classification on CIFAR-10 and CIFAR-100 and language modeling on WikiText-103, \iglu and \iglua achieve competitive or superior performance against \relu, \gelu, and other self-gated baselines, with the most pronounced gains appearing on heavily imbalanced datasets, where the heavy-tailed Cauchy gate provides a better distributional match to skewed input statistics.

\bibliographystyle{splncs04}
\bibliography{refs}

\end{document}